# Computer-Aided Diagnosis of Label-Free 3-D Optical Coherence Microscopy Images of Human Cervical Tissue

Yutao Ma, *Member, IEEE*, Tao Xu, Xiaolei Huang, Xiaofang Wang, Canyu Li, Jason Jerwick, Yuan Ning, Xianxu Zeng, Baojin Wang, Yihong Wang, Zhan Zhang, Xiaoan Zhang, and Chao Zhou, *Member, IEEE*

*Abstract*— **Objective:** Ultrahigh-resolution optical coherence microscopy (OCM) has recently demonstrated its potential for accurate diagnosis of human cervical diseases. One major challenge for clinical adoption, however, is the steep learning curve clinicians need to overcome to interpret OCM images. Developing an intelligent technique for computer-aided diagnosis (CADx) to accurately interpret OCM images will facilitate clinical adoption of the technology and improve patient care. *Methods:* 497 high-resolution 3-D OCM volumes (600 cross-sectional images each) were collected from 159 *ex vivo* specimens of 92 female patients. OCM image features were extracted using a convolutional neural network (CNN) model, concatenated with patient information (e.g., age, HPV results), and classified using a support vector machine classifier. Ten-fold cross-validations were utilized to test the performance of the CADx method in a five-class classification task and a binary classification task. *Results:* An 88.3±4.9% classification accuracy was achieved for five fine-grained classes of cervical tissue, namely normal, ectropion, low-grade and high-grade squamous intraepithelial lesions (LSIL and HSIL), and cancer. In the binary classification task (low-risk [normal, ectropion and LSIL] vs. high-risk [HSIL and cancer]), the CADx method achieved an area-under-the-curve (AUC) value of 0.959 with an 86.7±11.4% sensitivity and 93.5±3.8% specificity. *Conclusion:* The proposed deep-learning based CADx method outperformed three human experts. It was also able to identify morphological characteristics in OCM images that were consistent with histopathological interpretations. *Significance:* Label-free OCM imaging, combined with deep-learning based CADx methods, hold a great promise to be used in clinical settings for the effective screening and diagnosis of cervical diseases.

## I. INTRODUCTION

CERVICAL cancer is one of the most common cancers among women worldwide, especially in developing nations, and it has relatively high incidence and mortality rates [1]. Fortunately, cervical cancer is mostly preventable with active screening and detection techniques. For example, preventive screening and early detection can decrease the morbidity of cervical cancer by about 70% in the United States [2]. Nowadays, there are a few frequently-used cervical cancer screening techniques, such as high-risk human papillomavirus (HPV) testing, Pap smear cytology testing, colposcopy, and visual inspection of the cervix with acetic acid (VIA), each of which has its advantages and disadvantages.

Although HPV and Pap tests are widely used in women aged 25 and older to identify high-risk types of HPV that are most likely to cause cervical cancer [3] and abnormal cells, they cannot provide test results in real-time and are unable to localize cervical lesions. Instead, a VIA test allows clinicians to observe lesions and other changes in a patient's cervix directly, but it has lower sensitivity and specificity compared with HPV and Pap tests [4]. As the gold standard for diagnosing cervical cancer, colposcopy-directed biopsy with histopathological confirmation [5] is invasive and time-consuming and may cause complications to patients, such as bleeding, infection, and anxiety. Therefore, developing a non-invasive, efficient, and intelligent screening technique with relatively high sensitivity and specificity can significantly improve patient care.

Optical Coherence Tomography (OCT) [6] is an emerging biomedical imaging technique that utilizes light to obtain micrometer-resolution, cross-sectional images of biological tissue. By using high-resolution, high-speed OCT systems that can image cellular features of tissue samples up to 2 mm in depth in real-time [7], OCT has shown great potential as a non-invasive "optical biopsy" method [8], [9]. Previous studies have demonstrated the feasibility of using OCT for identification of morphological characteristics of the cervix, such as squamous epithelium, basement membrane, cervical stroma, low-grade and high-grade squamous intraepithelial lesions (LSIL and HSIL), and cervical cancers [10]–[15], which makes it possible to use OCT as a diagnostic tool adjunctive to colposcopy for cervical disease screening and detection [16].

This work was supported in part by the National Basic Research Program of China (Grant No. 2014CB340404), start-up fund from Lehigh University, US National Science Foundation (Grant No. DBI-1455613), US National Institutes of Health (Grant Nos. K99/R00-EB010071 and R01-EB025209), and Medical Science and Technology projects of China (Grant Nos. 201503117 and 161100311100). Yutao Ma and Tao Xu contributed equally to this work. (Corresponding author: Chao Zhou, e-mail: chaozhou@lehigh.edu).

Y. Ma and T. Xu are with the School of Computer Science, Wuhan University, China. X. Wang, C. Li, X. Zeng, B. Wang, Z. Zhang, and X. Zhang are with the Third Affiliated Hospital of Zhengzhou University, China. T. Xu, X. Wang, C. Li, Y. Ning, and X. Zeng were visiting students/scholars in the Department of Electrical and Computer Engineering, Lehigh University, USA. X. Huang is with the College of Information Sciences and Technology, Penn State University, USA. Y. Wang is with the Department of Pathology and Laboratory Medicine, Rhode Island Hospital/Warren Alpert Medical School of Brown University, USA. J. Jerwick and C. Zhou are with the Department of Bioengineering, Department of Electrical and Computer Engineering, Lehigh University, USA.

Optical Coherence Microscopy (OCM) [17], known as a combination of the coherent detection method of OCT and confocal microscopy, can provide better axial and lateral resolution than OCT. Moreover, ultrahigh-resolution OCM has recently shown the ability to reveal details of *ex vivo* cervical tissue similar to histology at the cellular level [18], significantly improving the diagnostic accuracy for cervical diseases. For example, in a recent blind test [18], three human experts achieved an average sensitivity of 80% (95% confidence interval, CI, 72%–86%) and an average specificity of 89% (95% CI, 84%–93%) for identifying high-risk lesions (including HSIL and invasive lesions) using label-free OCM images. The inter-observer agreement value of 0.627 suggests high diagnostic consistency among the three experts.

However, OCM images are foreign to gynecologists and pathologists due to the limited use of the OCM technology in clinics. Clinicians would need to undergo rigorous training, possibly involving viewing thousands of OCM images with different pathologies, to familiarize and recognize diagnostic features in OCM images. This training, however, can be time-consuming and may not be well-received by clinicians due to their busy clinical schedules. Since OCM images are different from those of traditional colposcopy and histological images, the training may create difficulty in clinical acceptance and adoption of this new technology for the screening and diagnosis of cervical diseases. Therefore, a computer-aided diagnosis (CADx) approach to effectively extract diagnostic imaging features and classify OCM images accurately would be highly desirable and can help facilitate broader adoption of the OCM imaging technology for clinical use.

In the past decade, deep learning [19] technologies utilizing deep neural networks have made remarkable progress in computer vision [20] and medical image analysis [21] for their capability to learn implicit or latent features from vast amounts of images and videos. More specifically, Convolutional Neural Networks (CNNs), a popular class of deep neural networks, have been widely used in image classification [22] and object detection [23]. Some recent studies suggest that deep CNNs can obtain results with accuracy comparable to and in some cases better than human experts for tasks such as image-based cancer (or rare disease) detection [24]–[27].

The purpose of this study is to develop a deep learning-based CADx method to evaluate cervical tissue samples using multi-modal feature information extracted from ultrahigh-resolution OCM imagery and routine medical exams such as the HPV test. We strive to accurately classify 3-D OCM images from *ex vivo* cervical specimens into "low-risk" and "high-risk" classes, which would pave the way for *in vivo*, real-time, and intelligent cervical diseases screening and diagnosis. Furthermore, we attempt to classify cervical lesions and provide their histopathological correlation with OCM imaging features in order to assist clinicians in understanding and interpreting OCM images.

TABLE I
STATISTICS OF THE EXPERIMENTAL DATASET

|  | Normal | Ectropion | LSIL | HSIL | Cancer | Total |
|---|---|---|---|---|---|---|
| Patients | 44 | 25 | 11 | 13 | 9 | 92 |
| Specimens | 71 | 32 | 16 | 18 | 22 | 159 |
| 3-D volumes | 197 | 79 | 28 | 55 | 138 | 497 |
| 2-D images | 55,070 | 23,150 | 7,714 | 16,500 | 39,033 | 141,467 |

## II. MATERIALS AND METHODS

### A. Data Collection

The experimental dataset used in this study contains 141,467 grayscale cervical tissue images collected from the Third Affiliated Hospital of Zhengzhou University, China, using a custom-developed ultrahigh-resolution OCM system [18]. Experimental protocol was approved by the Institutional Review Board of the Medical Faculty, Zhengzhou University. All the patients consented to the data collection. There were 159 cervical specimens from 92 female patients obtained from colposcopic biopsy ($n$ = 79), conization ($n$ = 26), and hysterectomy ($n$ = 54). All the 497 3-D OCM volumes for the specimens had a biopsy-confirmed diagnosis. Each OCM volume was annotated with a unique class label (see Section II.B). Table I describes the statistics of the experimental dataset. In addition to 3-D OCM volumes, patient information such as age and HPV test results were also collected, which have been clinically proved to be helpful when making a diagnosis for cervical lesions.

### B. Taxonomy of OCM Image Labels

The taxonomy of cervical OCM image labels consist mainly of five fine-grained classes: normal, ectropion, LSIL (CIN1), HSIL (CIN2&3), and cancer (including squamous cell carcinoma and adenocarcinoma). This taxonomy takes cervical ectropion (or cervical eversion) into consideration since it is often indistinguishable from early cervical cancer. However, cervical ectropion is a non-cancerous condition that occurs when the endocervix turns outward, exposing the columnar epithelium to the vaginal milieu [28]. As with our previous work [18], we also adopt two distinct general classes, i.e., "low risk" and "high risk." The general class "low risk" includes normal, ectropion, and LSIL, while the "high risk" class includes HSIL and cancer.

### C. Data Preparation

Each 3-D OCM image volume (~300 megabytes) contained a total of 600 2-D cross-sectional frames, each of which has 901 x 600 (height x width) pixels. The quality of OCM images played a vital role in training deep learning-based classification models (also called classifiers). We removed those 2-D cross-sectional frames that appeared to be "saturated" (i.e. too bright), too dark, or blurry from the original 3-D OCM volumes. According to Table I, the experimental dataset retained 141,467 high-quality cross-sectional 2-D OCM images. A 3 x 3 median filter was used to remove speckle noise from each 2-D OCM image. A single center crop with the size of 600 x 600 pixels was applied to each input image, and then the image was resized to 224 x 224 pixels while maintaining the original aspect ratio. Each 3-D OCM volume was zero-centered by subtracting the average intensity value from all the 2-D cross-sectional images within the volume.

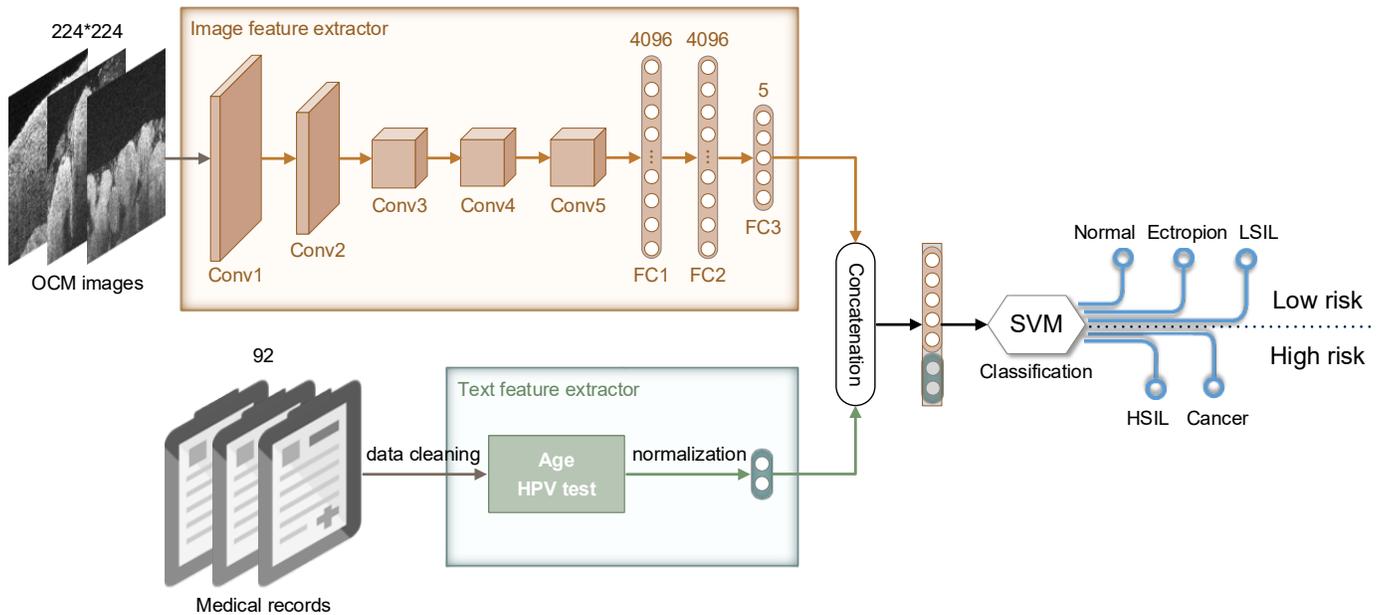

Fig. 1. The overall architecture of our CADx approach. First, we use VGG-16 to train an image feature extractor for ultrahigh-resolution OCM images. In particular, we append a new fully-connected layer (FC3) to the second fully-connected (FC2) layer. The FC3 layer outputs a 5-D feature vector representing the input image. Meanwhile, we build a text feature extractor to process patient information in medical records. Note that only age and HPV test results are available for this study. Second, we take advantage of a 7-D feature vector that concatenates the image and text features obtained to train an SVM-based classifier. Third, for a given OCM image, the SVM-based classifier outputs a predicted label of the five fine-grained classes. Besides, we evaluate the CADx method over two general classes, "low risk" (normal, ectropion, and LSIL) and "high risk" (HSIL and cancer), by inferring the probabilities of the corresponding fine-grained classes. The CADx method makes a decision of classification for each 3-D OCM image volume according to the mechanism of voting based on the majority rule.

Other patient data that are not images were processed using a text feature extractor. The patient age was converted from text and normalized using a min-max scaling, defined as follows.

$$x^* = \frac{x-min}{max-min}, \quad (1)$$

where $x$ denotes a patient's age, $min$ and $max$ represent the minimum value and the maximum value, respectively, of all patients' ages. Besides, HPV test results were defined by a Boolean datatype in this study. More specifically, "0" stands for a negative result, "1" for a positive result.

### D. Image Feature Extraction

We used VGG-16 (Visual Geometry Group 16-layer) [29], one of the most common CNNs, to train our image feature extractor. VGG-16 has 16 layers and it reduces the number of parameters in such a deep network using small (3 x 3) filters in convolutional layers. We modified the input image dimensions (224 x 224 x 3) for VGG-16 to ensure that the image feature extractor can process grayscale images (224 x 224 x 1). Also, instead of having the last fully-connected, softmax layer of VGG-16 output a vector of 1,000 categories, we replace that with a fully connected layer that outputs a vector of 5 categories (see FC3 in Fig. 1) to make it suitable for the specific five-class classification task based on the taxonomy defined in Section II.B. Another reason for adding the FC3 layer following the 4,096-D fully-connected layer (see FC2 in Fig. 1) was that our approach could achieve a better balance between the dimensionality of image features and the dimensionality of non-image features (i.e., age and HPV test results).

Because a few previous studies have reported that transfer learning using pre-trained models on the ImageNet dataset [30] is helpful to fine-tune CNN-based classifiers for medical grayscale images [31], [32], we also used pre-trained weights on a subset of the ImageNet dataset to fine-tune the image feature extractor for the sake of efficiency. In addition to the FC3 layer, we re-trained the first convolutional layer using OCM images and the last three convolutional layers (that is, conv5_1, conv5_2, and conv5_3) that capture task-specific features. All the hidden layers mentioned above were fine-tuned using the same global learning rate of 0.002. Moreover, we took advantage of the *Adam* (short for Adaptive Moment Estimation) optimization algorithm [33], with $β_1$ = 0.9, $β_2$ = 0.999, and a decay of 0.0002.

We trained, validated, and tested the image feature extractor using Keras (https://keras.io) and Google's TensorFlow (https://www.tensorflow.org) deep learning framework on a computer equipped with Intel Core i7 7200HQ processor, 64 GB RAM (Random-access Memory), and an NVIDIA GeForce GTX 1080 GPU (Graphics Processing Unit). The operating system was 64-bit Microsoft Windows 10.

### E. Classification Algorithm

Since support vector machines (SVMs) [34] can efficiently perform linear classification and non-linear classification tasks, they are helpful in text categorization and image classification. In this study, we developed an SVM-based classifier that utilized the multi-modal feature information (see Fig. 1). As stated above, we trained a CNN-based model to extract a 5-D feature vector from each OCM image. At the same time, we processed the age and HPV test results to obtain a 2-D non-image feature vector. The concatenation of the two types of features, which are 7-D feature vectors, are used as the input for the SVM-based classifier. The classifier outputs the probability of a given test sample (including an OCM image and its corresponding patient information) belonging to each of the five fine-grained classes. Note that the SVM-based classifier we utilized is based on an open-source tool called scikit-learn (http://scikit-learn.org), with default settings.

Recall that each 3-D OCM volume contains many 2-D cross sectional images, thus in order to obtain a probability distribution of a given 3-D OCM volume **y** over the five fine-grained classes, we calculate the probability using the following equation.

$$P(\mathbf{y} = j) = \frac{1}{|\mathbf{y}|} \sum_{i=1}^{|\mathbf{y}|} I(y_i = j), \quad (2)$$

where **y** denotes the whole OCM volume consisting of a set of 2D cross-sectional images, |**y**| is the total number of 2-D images, $j$ represents the class label ranging between 0 and 4, and $y_i$ is the class label predicted by the SVM-based classifier for the $i^{th}$ 2-D image included in **y**. Here, $I(x)$ is an indicator function defined as below.

$$I(x) = \begin{cases} 1 & \text{if } x \text{ is TRUE,} \\ 0 & \text{if } x \text{ is FALSE.} \end{cases} \quad (3)$$

After calculating the probability that a given 3-D OCM volume belongs to a specific fine-grained class, we then infer the likelihood that it belongs to a general class by summing up the probabilities over all the subclasses of the general class.

$$P(m) = \sum_{n \in S(m)} P(n), \quad (4)$$

where $m$ is a general class in the taxonomy mentioned above (i.e., "low risk" or "high risk"), $n$ is a fine-grained class, and the function $S(m)$ returns all the subclasses of $m$. Note that, in both fine-grained five-class and general two-class classification tasks, the SVM-based classifier made decisions according to the mechanism of voting based on the majority rule.

We evaluated the performance of the CADx method using ten-fold cross-validation. More specifically, we partitioned the experimental dataset into ten subsets of equal size, one of which was retained as the validation data for testing the CADx method and the other nine subsets were used as training data. This cross-validation process was repeated ten times, with each of the ten subsets used only once as the validation data, and the final evaluation was the average of the ten results. When partitioning experimental data into a training set and a test set for each iteration of the cross-validation process, all the OCM images from the same specimen were placed wholly into either the training set or the test set, to ensure that the two sets were mutually exclusive. This strategy helped prevent overfitting in the CADx model.

*F. Blind Test for Human Experts*

To compare the difference between human and machine in classification result, one experienced OCM researcher and two experienced pathologists from different institutes participated in this study. By using a three-step method including training, pre-testing, and blinded-testing [18], the three experts evaluated 297 3-D OCM volumes (i.e., a subset of the experimental dataset) separately and independently. Patient age and HPV results were provided to the human experts and taken into consideration when the diagnosis were made. Details of the three-step training and testing method and results were reported in [18].

*G. Evaluation Metrics*

*Accuracy* (or Trueness) is a descriptor of systematic errors, also known as a measure of statistical bias. *Sensitivity* (also called the true positive rate) measures the proportion of actual positives that are identified as such correctly, while *specificity* (also called the true negative rate) measures the percentage of actual negatives identified as such correctly. The following equations formulate their respective definitions in a binary classification task.

$$Accuracy = \frac{TP+TN}{TP+FP+FN+TN}, \quad (5)$$

$$Sensitivity = \frac{TP}{TP+FN}, \quad (6)$$

$$Speciicity = \frac{TN}{TN+FP}, \quad (7)$$

where *TP*, *TN*, *FP*, and *FN* represent the numbers of true positives, true negatives, false positives, and false negatives, respectively.

We used the estimated probability and actual class label to calculate the sensitivity and specificity of "low-risk" and "high-risk" diagnosis. In this study, a "low-risk" diagnosis includes normal, ectropion and LSIL, while a "high-risk" diagnosis includes HSIL and cancer. A 3-D OCM image is classified as "positive" if its estimated probability of being "high-risk" is larger than a given probability threshold $t$, and "negative" if the estimated probability is below the threshold. We then plotted the receiver operating characteristic (ROC) curve for the two general classes when varying the value of $t$ between 0 and 1. The area under the ROC curve (AUC) was also used to measure how well the CADx method performed in the binary classification task.

*H. Classification Performance Visualization*

A confusion matrix, also known as an error matrix [35], is used to visualize the performance of a classification algorithm. Each row of a confusion matrix denotes the instances (or class labels) of an actual class while each column indicates the cases of a predicted class, or vice versa. In this study, confusion matrices were used to show the misclassifications made by the CADx method and the three investigators on different (fine-grained or general) classes in two classification tasks. Each cell $(i, j)$ in these confusion matrices represents the empirical probability of predicting class $j$ given that the actual class label is $i$. The darker the color are off-diagonal elements, the higher the error rate. Therefore, a darker off-diagonal element indicated that the CADx method or the human expert under discussion had trouble distinguishing between the two given classes.

TABLE II
PERFORMANCE COMPARISON BETWEEN OUR CADx METHOD AND THREE HUMAN EXPERTS

| | Accuracy | | Sensitivity | | Specificity | |
|---|---|---|---|---|---|---|
| | Five classes | L/H | L/H | CI | L/H | CI |
| Investigator1 | 0.808 | 0.889 | 0.922 | 0.857~0.964 | 0.868 | 0.804~0.909 |
| Investigator2 | 0.700 | 0.845 | 0.817 | 0.735~0.883 | 0.863 | 0.810~0.914 |
| Investigator3 | 0.721 | 0.828 | 0.652 | 0.558~0.739 | 0.940 | 0.894~0.969 |
| Average (95% CI) | 0.743 | 0.854 | 0.797 | 0.717~0.862 | 0.890 | 0.836~0.931 |
| CADx method | 0.883(±0.049) | 0.913(±0.051) | 0.867(±0.114) | / | 0.935(±0.038) | / |

Five classes: normal, ectropion, LSIL, HSIL, and Cancer; L/H: low risk and high risk; CI: confidence interval. Confidence intervals for sensitivity and specificity are "exact" Clopper-Pearson confidence intervals [36] at the 95% confidence level. The three investigators' classification results are cited from [18].

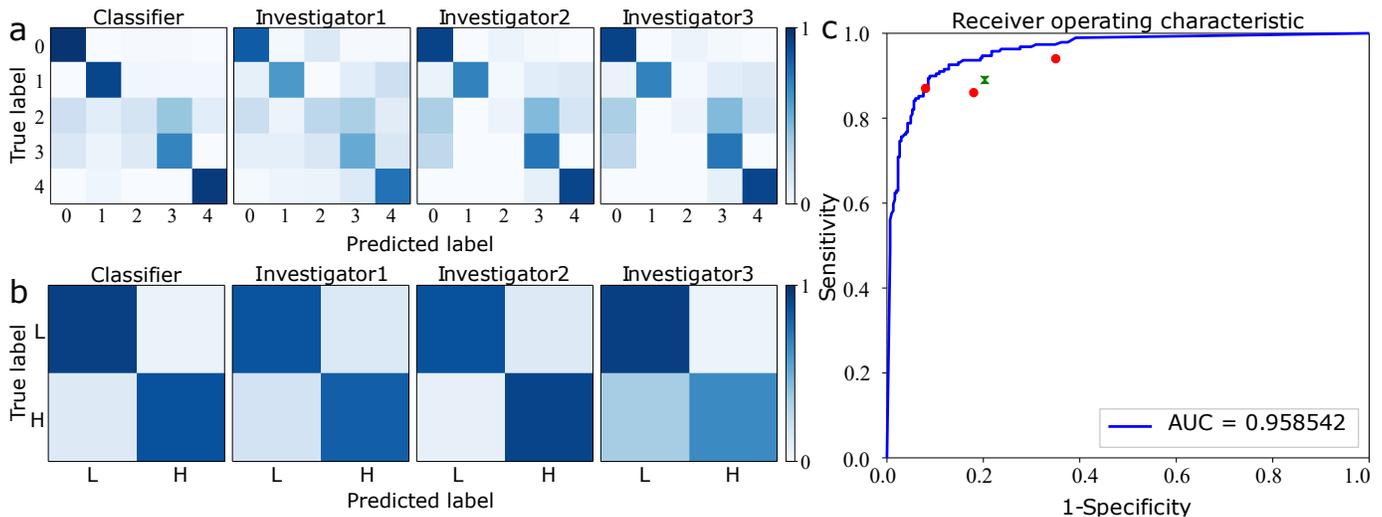

Fig. 2. Performance comparisons between the CADx method and three human experts. (a) Confusion matrices for the five-class classification task. "0," "1," "2," "3," and "4" represent normal, ectropion, LSIL, HSIL, and cancer, respectively. (b) Confusion matrices for the binary classification task. "L" and "H" represent low-risk and high-risk diagnosis, respectively. (c) ROC curve for the binary classification task. The X-axis stands for the false positive rate (1 − specificity), while the Y-axis represents the true positive rate (sensitivity). The (blue) ROC curve indicates the performance of the CADX method, each solid (red) circle denotes the performance of a human expert, and the (green) cross represents the average level of the three human experts.

*I. Image Feature Visualization*

We utilized two commonly-used feature maps to interpret how pixel-level image features "perceived" by the CNN-based image feature extractor differ among the five fine-grained classes. Springenberg *et al.* [37] proposed a method of guided backpropagation (GB). In this method, all the neurons (or called nodes in an artificial neural network) act like detectors of particular image features, thus explicitly visualizing diagnostic OCM image features for different classes. We also used a saliency map [38] to highlight visually dominant pixels based on the unique quality of each pixel, such as gray level intensity and image texture, etc..

### III. RESULTS

*A. Comparison between Human Experts and Machine on Five-Class classification*

As shown in Table II, in the five-class classification task, the CADx method achieved 88.3 ± 4.9% (mean ± s.d.) overall accuracy on the whole experimental dataset, while the three human experts obtained 80.8%, 70.0%, and 72.1% accuracy, respectively, on a subset of the experimental dataset, including 297 3-D OCM volumes.

In general, the three human experts and the machine learning classifier can differentiate normal cervix (see class 0 in Fig. 2a) from the other four cervical diseases. The CADx method only had a classification error rate of 2.0% for normal cervical tissue, comparable to that of the best human investigator (that was, 0.9%). Besides, it has less misclassification error in cervical cancer (see class 4 in Fig. 2a) than the three human experts. The classification error rates of the CADx method and the best human expert were 4.5% and 8.5%, respectively on cancer detection. Although cervical ectropion (see class 1 in Fig. 2a) is not an abnormality, the three human experts each misclassified a few 3-D OCM volumes of this class into LSIL (see class 2 in Fig. 2a), HSIL (see class 3 in Fig. 2a), or cancer, with classification error rates of 40.4%, 31.9%, and 36.2%, respectively. Instead, the CADx method's classification error rate for ectropion was only ~10.1%, much less than that of human experts. Because ~20% of misclassifications made by the three human experts occurred between ectropion and cancer [18], the above result indicated that the CADx method had a greater ability to distinguish between these two classes' irregular features in OCM images.

*B. Comparison between Human Experts and Machine on Binary Classification*

Fig. 2b presents the confusion matrices of the CADx method and the three human experts over the two general classes, showing that they can identify low-risk and high-risk 3-D OCM volumes with low misclassification rate. For example, the classification error rate of the CADx method for "low risk" was 6.9%, which was very close to that of the best human expert (that was, 6.0%). As shown in Table II, the CADx method achieved 91.3 ± 5.1% overall accuracy, while the three human experts obtained 88.9%, 84.5%, and 82.8% accuracy, respectively, in the binary classification task. In addition to its good classification performance, the CADx method was more consistent in diagnosis than human experts due to the differences among individual expert in skills and experience. For example, Fig. 2b shows that the second expert appeared to be good at discerning high-risk 3-D OCM images and tended to misclassify those uncertain test samples into "high risk," and the third expert appeared to be quite the opposite. Consequently, the former had a false positive rate of 13.2%, and the latter had a high false negative rate of 34.8%.

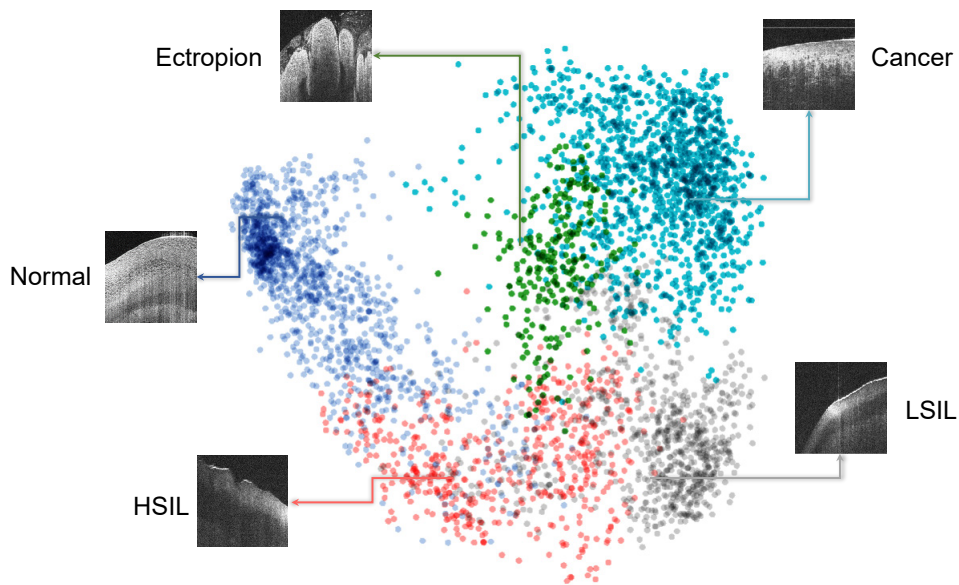

Fig. 3. Visualization of high-level representations of OCM images for the five fine-grained classes. Here, we illustrate 4,096-D feature vectors learned by the CNN-based image feature extractor for 2,500 randomly-selected OCM images in a Cartesian plane using PCA, known as a commonly-used dimension reduction method. Each colored point cloud represents a fine-grained class. Insets show images corresponding to various points in the point clouds. PCA embeds high-dimensional features in a low-dimensional space while preserving the pairwise distances of all the points that belong to different classes.

Fig. 2c depicts a ROC curve of the CADx method. It was evident from Fig. 2c that the CADx method outperformed two human experts who were denoted by solid (red) circles lying below the (blue) ROC curve, and it was equally valid as the remaining one expert in this binary classification task. As a result, the (green) cross, which represents the average of the three human experts, was also below the ROC curve. The AUC value of the SVM-based classifier reached 0.959, with a sensitivity of 86.7 ± 11.4% and a specificity of 93.5 ± 3.8%, indicating that it performed well overall and could provide diagnoses like experts.

Furthermore, unsurprisingly, the CADx method was far more efficient than human experts in both the binary and multi-class classification tasks. It took the CADx method ~4.8 milliseconds on average to complete diagnosis of a cross-sectional OCM image, and ~3 seconds to make a diagnosis of a 3-D OCM volume containing 600 2-D cross-sectional images. In contrast, it took a human expert several minutes to finish viewing a 3-D OCM volume and record a diagnosis.

*C. Visualization of High-Level Representations of OCM Images*

As shown in Fig. 1, the FC2 layer of the CNN-based image feature extractor outputs a 4,096-D feature vector for each input OCM image. We visualized this high-dimension feature of OCM images using principal component analysis (PCA) [39] in Fig. 3. Colored point clouds denote the five fine-grained classes, showing how the CADx method groups 2,500 randomly-selected OCM images into different classes (or clusters). OCM data in the same class are represented with the same color and clustered close to each other than to data from other classes. A 3-D view of the data clouds is shown in Video S1.

Generally speaking, the high-level representations obtained were semantic and had high intra-class similarity and apparent inter-class difference, which indeed facilitated the accurate classification of OCM images. Take the classification of cervical ectropion and cervical cancer samples as an example. Although it was not an easy task for the three investigators to differentiate cervical ectropion from cervical cancer in OCM image features (see the result introduced in Section III.A), the CADx method was able to detect distinct differences in the high-level feature space, with some minor overlaps between the two clusters (see Fig. 3). Thus, the CADx method only misclassified ~2.5% of cervical ectropion samples as cervical cancer and grouped ~4.5% cervical cancer samples into cervical ectropion incorrectly. This result was much better compared to those of the three human experts.

*D. Visualization of Pixel-Level Morphological Features in OCM Images*

Since deep learning is often considered as a "black box" [40], another major challenge in the development of the CADx method was to extract feature representations from OCM images and associate them with established morphological characteristics in histology. In Fig. 4, we utilized GB and saliency maps to visualize and highlight characteristic OCM image features for a normal cervical tissue, cervical ectropion, and cervical cancer. In addition to the original OCM image, we also presented the corresponding H&E (Hematoxylin and Eosin) histologic section to correlate features observed from OCM images.

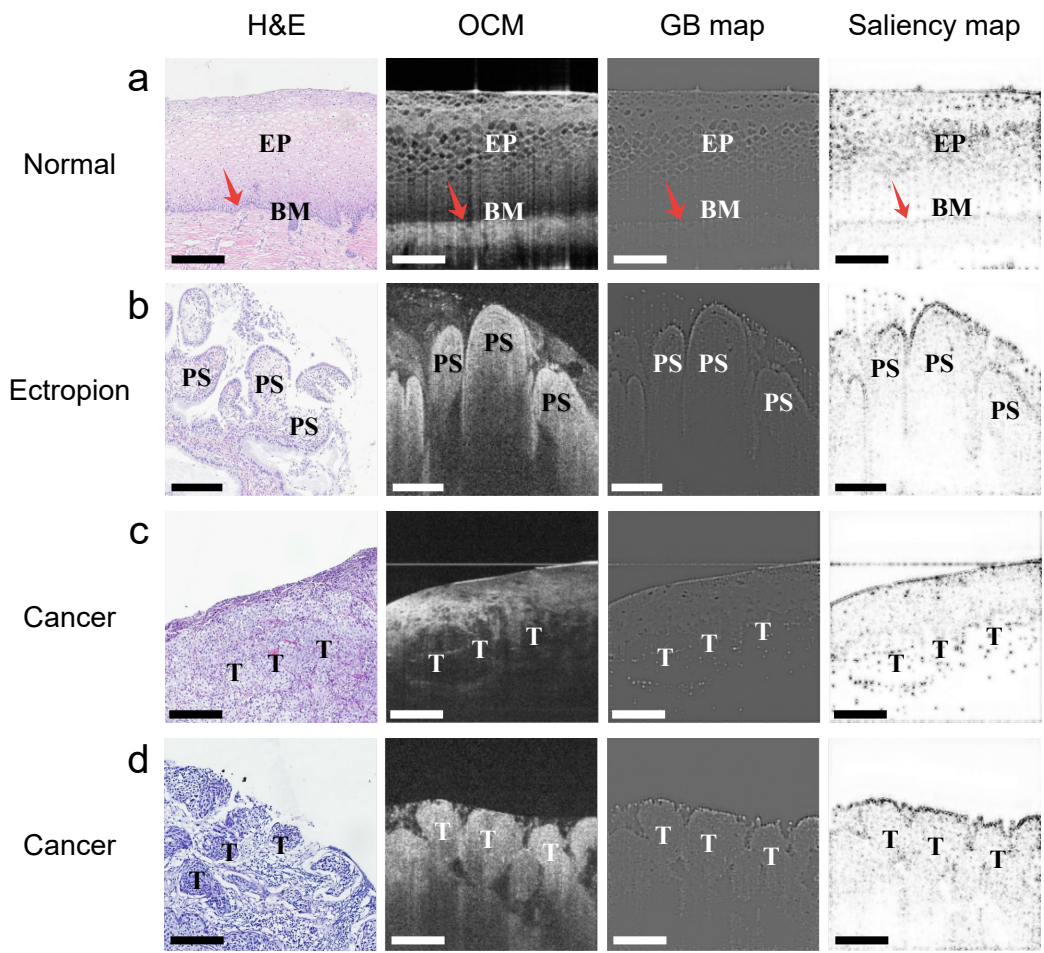

Fig. 4. Visualization of pixel-level morphological characteristics in OCM images extracted by our CADx method for three types of cervical tissue. (a) Normal tissue. (b) Cervical ectropion. (c&d) Cervical cancer. In each of the four plots, there are four different types of images, namely H&E histologic section, OCM image, GB map, and saliency map. GB and saliency maps highlight pixel-level morphological representations learned by the CNN-based image feature extractor. EP: squamous epithelium; BM: the basal membrane; PS: the papillary structure with interpapillary ridges; T: Tumor. Scale bars: 200 $\mu$m.

As shown in Fig. 4a, the OCM image of a normal cervical tissue sample showed a layered architecture of stratified squamous cells (EP) and the stroma, separated by the basement membrane (BM). These features matched well with the corresponding H&E image. Both the GB map and the saliency map highlighted the mesh-like epithelial cells and the layered structure with a clear and smooth interface between the basal layer of the epithelium and the stromal layer, which was one of the most striking morphological characteristics recognized in previous studies [11], [13], [18].

The OCM image in Fig. 4b demonstrated an example of cervical ectropion. The layered architecture of the epithelium was lost, and papillary structures with hyper-scattering boundaries were formed. The hyper-scattering boundaries of the papillary structures and interpapillary ridges were visualized in the OCM image as well as in the GB and saliency maps.

Fig. 4c and 4d present two examples of invasive cervical cancer (more specifically, cervical squamous cell carcinoma). In Fig. 4c, the GB and saliency maps captured a common morphological characteristic of cervical cancer. The epithelium became unstructured and disorganized, and the basement membrane was no longer observed, thus leading to a complete lack of architectural polarity. Besides, the two maps identified some sheets/nests of heterogeneous regions composed of epithelial cells and tumor cells. In Fig. 4d, the GB and saliency maps highlighted another common diagnostic feature of cervical cancer, that is, the microstructure of the cervical tumor sample disappeared, and oval-shaped clusters of homogeneous regions composed of nests of tumor cells were observed.

IV. DISCUSSION

In this paper, we applied a deep-learning based CADx method to diagnose cervical diseases based on label-free and non-destructive OCM images. The CADx method was trained to combine multi-modal feature information extracted from OCM imagery and patient information such as age and HPV test results to make a diagnosis. Moreover, it worked similarly to pathologists who often make a diagnosis based on the histology slides of biopsy specimens, taking into consideration patient information. Since it usually takes two or three days to get a pathology (or biopsy) report, we argue that the proposed OCM imaging and CADx method may simplify the workflow for cervical disease screening and generate diagnostic reports in a timely fashion.

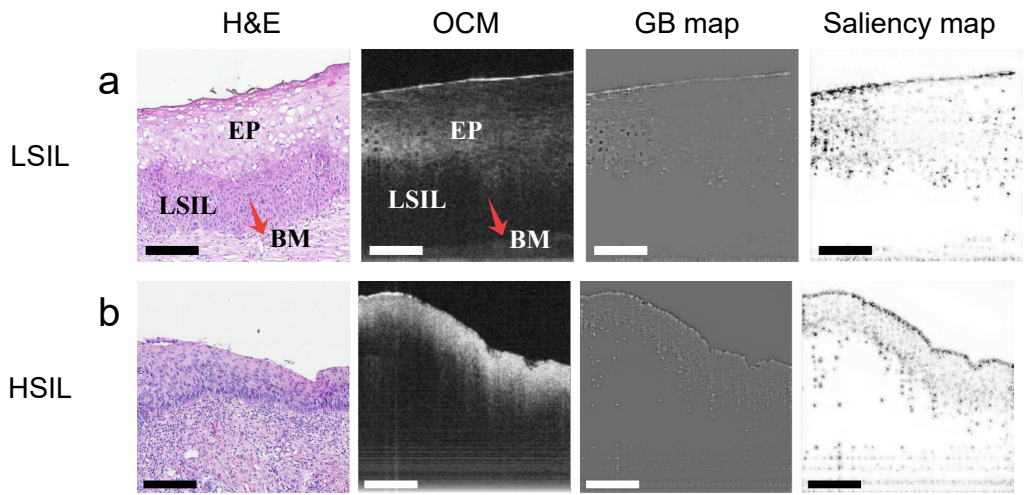

Fig. 5. Two confusing examples that were misclassified by our CADx method. (a) LSIL. (b) HSIL. In each of the two plots, there are four different types of images, namely H&E histologic section, OCM image, GB map, and saliency map. EP: squamous epithelium; BM: the basal membrane. Scale bars: 200 $\mu m$.

Previously, Kang *et al.* developed a CADx algorithm to classify OCT images of the human cervix using a linear discriminant analysis [41]. Nevertheless, it depended heavily on handcrafted image features, for example, the thickness of the epithelium and the contrast between the epithelium and the stroma, leading to a low average sensitivity of 51% (95% CI, 36%–67%) on a small-scale dataset including 152 cross-sectional OCT images. In this study, we validated the effectiveness of the CADx method in two classification tasks using ten-fold cross-validation. Compared with three human experts, the CADx method achieved higher overall accuracy (> 88%) in both the two-class and five-class classification tasks. Besides, in the binary classification task, the CADx method achieved an AUC value of 0.959 with 86.7 ± 11.4% sensitivity and 93.5 ± 3.8% specificity, showing its impressive performance in classifying 3-D OCM volumes of human cervical tissue. Compared with the traditional CADx algorithm [41], our method had a greater ability to effectively extract diagnostic imaging features from ultrahigh resolution OCM images.

The proposed CADx method still has room for improvement, especially in distinguishing between LSIL and HSIL. Even for expert pathologists, it was not an easy task to distinguish LSIL and HSIL from OCT images [12], [15]. Fig. 5 illustrates two examples misclassified by the CADx method. The low-grade lesion (see Fig. 5a) was incorrectly classified as HSIL, while the high-grade lesion (see Fig. 5b) was misclassified as LSIL. In Fig. 5a, the OCM image presented a layered architecture, including the superficial layer of the epithelium, a lesion of LSIL (at about the lower third of the epithelium), and the stroma. Moreover, the low-grade lesion displayed a hypo-scattering (poorer scattering) feature in the OCM image. According to

[15], [18], one of the most common optical feature of HSIL lesions on OCM was that more than one-third of the epithelium became irregular and thicker (for example, the OCM image in Fig. 5b). Sometimes, such a cervical lesion may span about two-thirds of the epithelium, and the basement membrane may not be visible. Unfortunately, the saliency and GB maps in Fig. 5 showed that the CNN-based image feature extractor, which was trained with limited samples of LSIL and HSIL, failed to capture the morphological features mentioned above that are meaningful to human experts. As a result, the CADx method produced similar probabilities for LSIL and HSIL. Hence, our CADx method had a limited ability to differentiate HSIL from LSIL (also see the overlap between the clusters of LSIL and HSIL in Fig. 3) at the moment due to limited samples (e.g., only 28 LSIL and 55 HSIL 3-D OCM image volumes were included in the current study).

Our future work is to tackle the above problem from two aspects. On the one hand, we will collect more samples of LSIL and HSIL *ex vivo* to train a better image feature extractor. Furthermore, we will take advantage of *en face* OCM images obtained from the 3-D OCM volume when training the image feature extractor. Our previous study has showed that cervical ectropion had very different image features compared to invasive lesions in an *en face* OCM image [18]. On the other hand, we plan to enhance the learning ability of the image feature extractor using deep reinforcement learning [42] in combination with human knowledge and skills as well as instant feedback.

## V. Conclusion

In summary, we developed a deep-learning based CADx method and applied it to diagnose cervical diseases based on 3-D OCM image volumes and patient information. The CADx method was shown to be effective in binary and multi-class classification tasks, demonstrating classification accuracies better than three human experts. Using guided backpropagation and saliency maps, we further identified morphological characteristics in OCM imagery to provide histopathological correlation of OCM image features. With the assistance of the CADx method, ultrahigh resolution OCM technology holds the potential to become a promising complement to existing technologies for noninvasive, label-free and real-time screening and diagnosis of human cervical diseases.

ACKNOWLEDGMENT

We thank Wei Zhang, Yiming Du, Di Meng, Xian Cui and Linyu Li for assistance in collecting and imaging the specimens.